\documentclass[english]{amsart}
\usepackage[T1]{fontenc}
\usepackage[latin9]{inputenc}
\usepackage{xcolor}
\usepackage{units}
\usepackage{amsthm}
\usepackage{amsmath}
\usepackage{esint}
\usepackage{longtable}
\usepackage{tabu}
\PassOptionsToPackage{normalem}{ulem}
\usepackage{ulem} 
\usepackage{natbib}

\makeatletter

\providecolor{lyxadded}{rgb}{0,0,1}
\providecolor{lyxdeleted}{rgb}{1,0,0}

  \theoremstyle{plain}
  \newtheorem*{thm*}{\protect\theoremname}
  \newtheorem*{example*}{\protect\examplename}

\usepackage[all]{xy}

\usepackage{placeins}

\usepackage{lineno}

\newcommand{\Sup}{\mathsf{Sup}}\newcommand{\Tol}{\mathsf{Tol}}\newcommand{\Con}{\mathsf{Con}}\newcommand{\T}{\mathsf{True}}\newcommand{\F}{\mathsf{False}}\newcommand{\TV}{\mathsf{TV}}\newcommand{\Prob}{\mathsf{Prob}}\newcommand{\dd}{\mathsf{d}}

\def\frontmatter@abstractheading{}

\makeatother

\usepackage{babel}
  \providecommand{\theoremname}{Theorem}
  \providecommand{\examplename}{Theorem}

\begin{document}

\title{Classificatory Sorites, Probabilistic Supervenience, and Rule-Making}

\author{Damir D. Dzhafarov}
\address[D.~D.~Dzhafarov]{Department of Mathematics\\ University of Connecticut\\
196 Auditorium Road\\ Storrs, Connecticut 06269 U.S.A.}
\email{damir@math.uconn.edu}

\author{Ehtibar N. Dzhafarov}
\address[E.~N.~Dzhafarov, corresponding author]{Department of Psychological Sciences\\ Purdue University\\
703 Third Street\\ West Lafayette, Indiana 47907 U.S.A.}
\email{ehtibar@purdue.edu}

\thanks{\noindent The first author was supported by an NSF Postdoctoral Fellowship and NSF grant DMS-1400267, and the second author was supported by NSF grant SES-1155956.}

\maketitle
\begin{abstract}
We view sorites in terms of stimuli acting upon a system and evoking this system's responses. Supervenience of responses on stimuli implies that they either lack tolerance (i.e., they change in every vicinity of some of the stimuli), or stimuli are not always connectable by finite chains of stimuli in which successive members are `very similar'. If supervenience does not hold, the properties of tolerance and connectedness cannot be formulated and therefore soritical sequences cannot be constructed. We hypothesize that supervenience in empirical systems (such as people answering questions) is fundamentally probabilistic. The supervenience of probabilities of responses on stimuli is stable, in the sense that `higher-order' probability distributions can always be reduced to `ordinary' ones. In making rules about which stimuli ought to correspond to which responses, the main characterization of choices in soritical situations is their arbitrariness. We argue that arbitrariness poses no problems for classical logic.
\end{abstract}

\section{Introduction}

\subsection{Overview}

The purpose of this paper is to discuss and elaborate some aspects
of what we have called the \emph{behavioral approach} to sorites \cite{DD-2010a, DD-2010b}. Here, the word `behavior' is, perhaps, somewhat misleading, as we understand it in a broader way than is usual: namely, as any input-output relation in any system, not necessarily sentient
or biological. The central feature of this approach is that instead
of being concerned with whether a certain object $x$ has a certain
property $P$ `in reality', we deal with the question
of whether a system consistently responds to $x$ in a particular
way (which we then interpret as the system assigning a certain property $P$
to $x$). 

Consider an example. Aliya has to choose between four answers in response
to being shown an object $x$ (say a formation of grains of sand): 
\begin{longtable}{rll}
$+$ & : & `$x$ is $P$',\\
$-$ & : & `$x$ is $\overline{P}$',\\
$\pm$ & : & `$x$ is $P$ and $\overline{P}$',\\
$\cdot$ & : & `$x$ is neither $P$ nor $\overline{P}$'.
\end{longtable}
\noindent Here, $P$ is some property (say, `a heap') and $\overline{P}$
is its internal negation (`something other than a heap'). They are assumed
to be understood by Aliya, although we do not know exactly how. In
the behavioral approach we need not worry exactly how, insofar as
Aliya follows the rules of responding we impose on her. Thus, she
could have answered in many other ways, but we constrain her to choosing
between these particular four responses by the rules of responding.

\subsection{Traditional versus behavioral approach}

A traditional philosophical analysis would begin by translating the
four responses to $x$ into logical predicates 
\begin{longtable}{lll}
$P_{+}^{*}(x)$,\\
$P_{-}^{*}(x)$ & $=$ & $\overline{P^{*}}(x)$,\\
$P_{\pm}^{*}(x)$ & $=$ & $P^{*}(x)\wedge\overline{P^{*}}(x)$,\\
$P_{\cdot}^{*}(x)$ & $=$ & $\sim P^{*}(x)\wedge\sim\overline{P^{*}}(x)$.
\end{longtable}
\noindent Then the analysis would be directed to finding out whether the statement
`$P_{+}^{*}(x)$' is true or false (or anything else, if
one allows for non-classical logics); or what its truth value ought to be assuming the truth value of `$P_{-}^{*}(x)$' is specified; or whether it is
possible that the statement `$P_{\pm}^{*}(x)$' is true,
etc. The analysis may lead to distinguishing between different sorts
of $P$s. Thus, compared to precise predicates, such as `has amplitude
$A$ at wavelength $w$', one may declare the vague ones, like `is a heap', to have different logical relations with objects they apply
to and with other predicates. 

In the behavioral approach we treat $x$ as an input acting upon the
system (in this example, Aliya). Borrowing terminology from psychology,
$x$ may also be generically referred to as a \emph{stimulus}. Then $+$, $-$, $\pm$, and $\cdot$ are four possible values of the system's output, or \emph{response}. Our standing assumption here (elaborated upon below) shall be that responses supervene on stimuli, i.e., that responses are given consistently. In precise terms, this means that every instance of a given stimulus $x$ is associated with one and the same
response $r$, so that there is a function
$\pi$ (from the set of all stimuli to the set of responses) such that 
\[
r=\pi(x).
\]
In our example, the relation $\pi(x)=r$ for any $r \in \{ +,-,\pm,\cdot\} $
is interpreted as the fact that Aliya consistently assigns response
$r$ to $x$. One may then introduce a predicate $P_r (x)$ that holds if
and only if $\pi(x)=r$.

Let us compare the predicates $P_r$ to the predicates
$P_r^{*}$ of the traditional analysis. The predicates $P_r^{*}$ may very well
be characterized as vague, and their theory as glutty, gappy, or otherwise non-classical, but the predicates $P_r$
are always well-defined, and for each $x$, the statements `$P_r(x)$'
have definite classical truth values. In particular, the predicates $P_r$ are mutually exclusive and, assuming each stimulus is associated with a response, mutually exhaustive. There is nothing vague about Aliya's maintaining that $x$ has a certain vague property, nor even about her
maintaining that $x$ has a classically contradictory property (say, being both red and not red). In both cases she definitely assigns this response or definitely does not assign it to a given $x$. In other words, while the logic or objective truth of the intended meanings of Aliya's responses (of the predicates $P_r^*$) may in principle be arbitrary, the supervenience assumption binds Aliya's \emph{assignments} of responses (the predicates $P_r$) to classical logic. In the present example, if $\pi(x) = \pm$ then we infer that Aliya consistently assigns to $x$ being a heap and being something other than a heap. However, as $+$ and $-$ are different responses from $\pm$, and $\pi(x)$ does not equal either of the two, we infer neither that Aliya consistently assigns to $x$ being a heap, nor that she consistently assigns to $x$ being something other than a heap. (In fact, we infer the external negations of both these statements.) The disquotational
principle applies here in its clearest form: the
statement `Aliya consistently assigns response $r$ to $x$'
is true if Aliya consistently assigns response $r$ to $x$, i.e., if $P_r(x)$; the
statement is false if she does not, i.e., if $\sim P_r(x)$.
Thus, we may conveniently and innocuously confuse the predicates $P_r(x)$ with the statements `$P_r(x)$'. By contrast, confusing $P_r^{*}(x)$ with `$P_r^{*}(x)$'
may be conceptually more demanding, as it requires that one think
of the objective reality of things like `a heap' \citep{Dummett-1975,Unger-1979,Wheeler-1979}. 

\subsection{The soritical trap}

The reason we can get away with the behavioral approach in dealing
with sorites is that soritical reasoning can always be formulated
in terms of how a system's consistent responses to different objects
differ depending on how these objects differ from each other. Specifically,
all forms of the classificatory sorites (as opposed to the comparative sorites; see our conclusion for the difference
between these two forms) are pivoted
on the proposition that one and the same response $r$ should be given
to stimuli $x$ and $y$ that are \emph{maximally} or \emph{sufficiently} close,
where this closeness is understood in some objective sense, external and extraneous
to the responding system. Thus, Aliya may be asked to theorize about how
she would respond to a sand formation $y$ which differs by only
one grain of sand from another formation $x$, provided she has responded to $x$ by
$r$. Aliya may be tempted to declare that she will not change
her response because the difference by one grain of sand is too small
to make a difference. If she does, she will fall into the standard soritical
trap, and we will be able to construct a chain (\emph{soritical sequence}) $x_1,x_2,x_3,\ldots,x_n$
in which every two successive elements differ by one grain of sand,
and so by her own reasoning elicit the same response from Aliya, yet $x_1$ and $x_n$ differ by so many grains of sand (i.e., $n$ is so large) that Aliya responds differently to the two.

The supervenience assumption is not explicit in the soritical trap just described. But if supervenience is violated, i.e., if the function $\pi$ is not well-defined, the soritical trap cannot even be formulated, let alone `sprung'. Indeed, if Aliya has agreed that she would
respond to a stimulus $y$ by $r$ provided she responded to
a very similar $x$ by $r$, she should certainly agree to do the same for $y=x$. After all, nothing is more similar to
$x$ than another instance of $x$, under any reasonable definition of similarity. The very possibility that one and the same formation of sand may be a heap in one instance and not be a heap in another essentially deprives the classical sorites argument of that which makes it the most compelling. Supervenience is therefore an integral assumption, not merely a construct of the behavioral approach, as its rejection means to end the discussion of sorites right away. In point of fact, this may not be unreasonable given our common experience that individuals indeed can and do change their responses over time. Aliya, being human, may change her responses based on any number of factors, from the time of day to her (presumably waning) interest in answering questions about sand. She may even choose to answer randomly. But in such situations, as we argue below, supervenience can be seamlessly reinstated by extending $\pi$ from individual responses themselves to their probabilities. The `crux' of dealing with the soritical trap must thus lie elsewhere.

A technical complication here is that the value of $\pi(x)$ for a given $x$ cannot be established by a direct observation of what Aliya says in response to being presented with the stimulus $x$. This value is a theoretical assumption that can be corroborated, though
not proved, by observing her responses to repeated instances of the same stimulus, $x$. Experimentally, such instances can be created either by repeated presentations of $x$
to Aliya under fixed or well-counterbalanced conditions, or they can
be created by observing the responses to $x$ by many people considered
to be similar to Aliya (in relevant respects). Rather than getting into various designs of such corroborating observations (they can be found in any textbook of experimental design in behavioral and social sciences), we utilize the fact that this is
a philosophy paper and move to a more abstract plane of analysis. Whatever method is used to collect this information, we can assume for our purposes here to have access to it, as if from an infinity of parallel worlds
in which we observed an infinity of Aliyas decide about how to respond to $x$. If supervenience holds, be it deterministic or probabilistic, then the value of $\pi(x)$ can be obtained from the said information.

\subsection{Plan}

We will discuss below how the soritical trap is dissolved both on
the level of responding to stimuli (descriptive analysis) and on the
level of making rules about responding to stimuli (normative analysis).
To repeat, everything in this discussion can be expressed entirely in terms of what Aliya says
or thinks she would or should say in response to different sand formations, not about which of these do or do not make a heap in reality. However, the behavioral approach and the traditional one can be related
through the assumption of a `competent responder'. In our example,
if the predicates $P_r^{*}(x)$ are assumed to have objective
truth values (in the classical sense), then $P(x)$ can
be requested by the rules of responding to have the same truth values,
provided Aliya has all the relevant information about $x$ and can
compute $P_r^{*}(x)$. This means essentially that if
Aliya is competent and honest, then it will be the case that $P_r^{*}(x)$ holds if
and only if $P_r(x)$ does. In particular, if a soritical
sequence can be formed in terms of $P_r^{*}$, it will then also be
formable in terms of $P_r$. But precisely because $P_r$ is squarely
within classical logic, it admits no soritical sequences, whence neither does $P^{*}_r$.

The plan of the paper is as follows. In Section \ref{sec:Definitions} we outline the technical components of our approach, and formally derive the impossibility of the existence of soritical sequences. In Section \ref{sec:det}, we illustrate the dissolution of the paradox for systems where responses to stimuli are assumed to be deterministic, and in Section \ref{sec:prob} we do the same for probabilistic systems. In Section \ref{sec:obj} we delve deeper into the probabilistic model, and address a couple of natural concerns, including the tempting but misguided idea to generalize it to an increasing hierarchy of probabilities, each governed by the next. Finally, in Section \ref{sec:rules}, we address arbitrariness and justifiability in connection with normative rules for responding to stimuli.

\section{Basic Notions}\label{sec:Definitions}

\subsection{Systems}

To describe the soritical trap in formal terms, we begin by defining a \emph{system} $\mathcal{S}$ to be a structure $(S,R,\pi)$ in which $S$ and $R$ are sets and $\pi$ is a function $S \to R$. We interpret these components as follows:
\begin{itemize}
\item $S$ is a set of inputs to which the system responds, generically referred to as \emph{stimuli} (but sometimes also as \emph{objects}, \emph{points}, etc., depending on context);
\item $R$ is a set of outputs, generically referred to as \emph{responses} or \emph{stimulus-effects};
\item $\pi$ is called a \emph{response} or \emph{stimulus-effect} function, and maps stimuli to their (consistent) responses under the system.
\end{itemize}
To use the example from the introduction, $S$ can be the set of all possible formations of sand in the world, and $R$ the set $\{+,-,\pm,\cdot\}$. The assumption of the existence of $\pi$ is our supervenience assumption, and we denote it $\Sup$ here. As noted above, $\Sup$ is an implicit but fundamental piece of the soritical trap.

The generality of the above setup is apparent when one considers that each system can give rise to many others, which may be more natural or useful in different situations. For example, given a system $\mathcal{S} = (S,R,\pi)$, we can consider instead a system $\mathcal{S}'$ whose set of stimuli consists of finite sequences of elements of $S$. One might prefer to work with this system if the response to a given stimulus $x$ is believed to depend not only on $x$ itself, but also on the sequence of previously-seen stimuli \citep[as advocated, e.g., by][]{Raffman-2014}.

\subsection{Frech\'{e}t spaces}

We need next to formulate a notion of closeness between two stimuli, a precise (and practical) generalization of when two formations of sand differ by very few grains of sand. Contrary to one's intuition, this requires neither the imposition of a metric on $S$
\citep{Williamson-1994}, nor even of a topology \citep{WC-2010}. Our formalism \citep{DD-2010a, DD-2010b} is pre-topological, and almost certainly as general as possible for this notion. A set $S$ is said to be endowed with \emph{Fr\'{e}chet
vicinities} (and to form, together with them, a \emph{Fr\'{e}chet space}) if every
$x \in S$ is associated with a nonempty collection $\mathbf{V}_x$ of subsets of $S$
containing $x$ (our definition here is more restrictive than
in \cite{Sierpinski-1952}). The members of $\mathbf{V}_x$ are called the Fr\'{e}chet vicinities \emph{of} $x$, and closeness can be defined in terms of them thus:\\

\noindent \begin{tabu} to \linewidth {rX}
& $y\in S$ is close to $x\in S$ \emph{in the sense of} the Fr\'{e}chet vicinity $V \in \mathbf{V}_x$ if $y\in V$.
\end{tabu}\\

\noindent A point $x$ in a general Fr\'{e}chet space may have one, several, or infinitely many Fr\'{e}chet vicinities, and a point $y$ may be close to $x$ in the sense of all, some, or none of these. Importantly, $x$ is close to itself in all possible senses (as it belongs to each of its Fr\'{e}chet vicinities, by definition). In our example, we may choose to let $\mathbf{V}_x$ for each sand formation $x$ have just one Fr\'{e}chet vicinity, namely the set of all sand formations that can be obtained from $x$ by adding or removing one or fewer grains of sand. This turns the set $S$ of sand formations into a Fr\'{e}chet space, and the only sense in which two different formations can be considered close is if they differ by a single grain. In the general case, unlike here, closeness need not be symmetric: $y$ can be close to $x$ without $x$ being close to $y$.

\subsection{Tolerance}

In the abstract, the definition of a Fr\'{e}chet space is completely independent of our notion of a system. We connect the two with the following tolerance assumption, which we denote by $\Tol$:\\

\noindent \begin{tabu} to \linewidth {rX}
& if $\mathcal{S} = (S,R,\pi)$ is a system and $S$ is endowed with Fr\'{e}chet vicinities, then $\pi$ is \emph{tolerant}: it is constant on at least one Fr\'{e}chet vicinity of each $x \in S$.
\end{tabu}\\

\noindent To agree that the function $\pi$ is tolerant is the main (explicitly stated, unlike $\Sup$)
part of the classical soritical trap. For example, with vicinities assigned to sand formations as above, the stimulus-effect function $\pi$ corresponding to Aliya's assignment of responses will, in view of her commitment to respond the same way to any two sand formations that differ by a single grain of sand, satisfy $\Tol$: it will be constant on each $x$'s unique Fr\'{e}chet vicinity.

\subsection{Connectedness}

Finally, we need the vicinities to be connected in a particular way, an abstract way of going from something that is not a heap to something that is. A \emph{$V$-cover of $S$} is a collection $\mathbf{C}$ of subsets of $S$ each of which is a Fr\'{e}chet vicinity of some $x \in S$, and containing at least one Fr\'{e}chet vicinity of each such $x$. Two stimuli $x,y\in S$ can then be defined to be
\emph{$V$-connected} if from every $V$-cover of $S$ one can choose Fr\'{e}chet vicinities
$V_1,V_2,\ldots,V_n$ such that $x \in V_1$,
$y \in V_n$, and $V_i\cap V_{i+1} \neq \emptyset$
for all $i=1,\ldots,n-1$. To formulate a soritical trap one needs the
following connectedness assumption, denoted $\Con$:\\

\noindent \begin{tabu} to \linewidth {rX}
& there are at least two V-connected stimuli
$x,y\in S$ such that $\pi(x)\not=\pi(y)$.
\end{tabu}\\

\subsection{Putting it all together}

\noindent It is easy to prove now, as in \cite{DD-2010a}, that $\Sup$,
$\Tol$, and $\Con$ allow one to construct a classificatory soritical
sequence, i.e., $x_1,\ldots,x_k$ such that $\pi(x_i)=\pi(x_{i+1})$
for all $i=1,\ldots,n-1$ but $\pi(x_1)\not=\pi(x_k)$.
By reductio, $\Tol$ and $\Con$ cannot hold jointly for any function
$\pi$. Both these assumptions have to be made in order for the existence
of soritical sequences to be guaranteed: dropping either of them makes
the system logically consistent. The assumption $\Sup$ cannot be
dropped alone, as without it $\Tol$ and $\Con$ cannot be formulated.

\section{Deterministic Supervenience}\label{sec:det}

\subsection{Consistent responses}

Let us illustrate the consequences of this analysis with another example.
Max is being presented real numbers between $0$ and $1$ (inclusive)
and asked to classify them as `close to 1' (response $r_1$)
or `not close to 1' (response $r_0$). Thus, the set $S$ of stimuli here
is the real closed unit interval $[0,1]$, and the set $R$ of responses is $\{ r_1,r_0\} $. 
The Fr\'{e}chet vicinities of $x\in[0,1]$  are defined in a conventional way, e.g., as $(x-\varepsilon,x+\varepsilon)\cap[0,1]$ 
for all possible  $\varepsilon>0$. Any two $x,y$ in $[0,1]$ therefore are V-connected.

Suppose first that Max's responses are consistent, i.e., his choice
of $r\in R$ is uniquely determined by the number $x \in S$ presented:
$r=\pi(x)$. (That is, in all possible worlds Max's copies
always give the same responses to the same stimuli.) Assume further
the following rules of responding:
\begin{enumerate}
\item[(M1)] there are $x_0,x_1 \in S = [0,1]$ such that $\pi(x_0)=r_0$
and $\pi(x_1)=r_1$;
\item[(M2)] if $\pi(x)=r_1$, then $\pi(y)=r_1$
for all $y \geq x$ in $S$;
\item[(M3)] if $\pi(x)=r_0$, then $\pi(y)=r_0$
for all $y\leq x$ in $S$.
\end{enumerate}
It immediately follows from these rules that there should exist some
$v\in[0,1]$ such that either $\pi(x)=r_1$
if and only if $x\in[v,1]$, or else $\pi(x)=r_1$
if and only if $x\in(v,1]$. What is more, the value of this $v$ can be (empirically) estimated to any desired
degree of precision, e.g., by the following simple recursive algorithm, which produces a sequence of rational numbers $q_0,q_1,\ldots$ such that $q_n$ is within $2^{-n}$ of the value of $v$: let $q_0 = 0$, and given $q_n$ for some $n \geq 0$, let $q_{n+1} = q_n$ if $r(q_n + 2^{-(n+1)}) = r_1$, and let $q_{n+1} = q_n + 2^{-(n+1)}$ otherwise. Unless $v$ happens to be rational, this is as precise a method of specifying the value of a real number as possible. We can thus legitimately claim to `know' this value, at least insofar as we can know the value of most real numbers: we know it in precisely the same way we know the value of $e$ or $\sqrt{2}$. We therefore cannot see how
one can accept that $\pi(x)$ exists and follows
Max's rules, so that the existence of $v$ follows, without also accepting the knowability of this $v$. But this seems to be the epistemic position on sorites, cf.~\cite{Sorensen-1988a, Sorensen-1988b}; \cite{Williamson-1994, Williamson-1997, Williamson-2000}.

\subsection{A rational response}

Consider the possibility of trapping a rational person into the
soritical paradox using Max's rules. The rational person's name
is Alex, and we present the situation as a conversation between her
and Eubulides. First, Eubulides describes Max's rules to Alex and
asks her to accept them as given. Then he proceeds.
\begin{longtabu} to \linewidth {rX}
Eubulides: &  Will you agree that 1 is close to 1?\\
Alex: &  Yes, because it follows from rules M1 and M2.\\
Eubulides: &  Will you agree that,\\
& \begin{tabu}{rX}
(E1) & if $x$ is close to 1 and $\varepsilon>0$ is sufficiently small, then $x-\varepsilon$ is also close to 1? 
\end{tabu}\\
& And, for symmetry, will you agree that\\
& \begin{tabu} to \linewidth {rX}
(E2) & if $x$ is not close to 1 and $\varepsilon>0$ is sufficiently small, then $x+\varepsilon$ is also not close to 1?
\end{tabu}\\
Alex: &  Let me see. I know that Max's rules require the existence of a certain number $v$ such that his responses are determined by one of two rules:\\
& \begin{tabu} to \linewidth {rX}
(1) & $x$ is close to 1 if and only if $x\geq v$, or\\
(2) & $x$ is close to 1 if and only if $x>v$.
\end{tabu}\\
& In the first case I agree with E2, because for any $x<v$,
I can always choose an $\varepsilon>0$ so that $x+\varepsilon<v$.
But I have to reject E1, because the statement does not hold for $x=v$.
In the second case, by analogous reasoning, I agree with E1 but have
to reject E2.\\
Eubulides: &  We can take the first case without loss of generality, so you agree with E2. Does it not lead to an increasing sequence in which
the first number is not close to 1, and the second is also
not close to 1, and the third, and so
on? And yet since the numbers get larger and larger, will you not eventually reach a value large enough to be close
to 1?\\
Alex: &  Not at all. My acceptance of E2 allows for the $\varepsilon$ to depend upon the $x$. For every given $x < v$, I thus choose a positive $\varepsilon=\varepsilon(x)<v-x$. Then the
(infinite!) sequence\\
& \center \begin{tabu}{l}
$x$,\\
$x + \varepsilon(x)$,\\
$x + \varepsilon(x) + \varepsilon(x+\varepsilon(x))$,\\
$\vdots$
\end{tabu}\\
& can never exceed $v$, so I will deem no number occurring in it as being close to $1$.\\
Eubulides: &  But what is this number $v$? Can you find out its
value? \\
Alex: & Provided Max follows his rules (and you told me he did),
then I know such a $v$ must exist, just as $\sqrt{2}$ exists. As for its value, it is whatever it is, and if I can ask Max questions, I can approximate it as accurately as you wish me to.
\end{longtabu}

\noindent Put formally, although the system does satisfy $\Con$ and although $\Tol$ is consistent with Max's rules for all $x<v$, $\Tol$ does not hold for $x=v$, and so Eubulides must realize he cannot set a soritical trap for Alex. Another way of formalizing the situation would be to endow $S=[0,1]$ with unconventional vicinities, e.g., of the form $[x,x+\varepsilon)\cap[0,1]$ for every $x$: in this case no two distinct points in $[0,1]$ are V-connected, and a soritical sequence cannot be constructed even if $\Tol$ is agreed to hold throughout $[0,1]$ (which in this case is equivalent to accepting Eubulides' E2).

\subsection{Conclusion} Why do we not normally see the situation as clearly
as Alex does? Why are we so easily trapped into agreeing that if someone
is bald, then adding a single hair would leave him still bald? It
should be clear that if some heads are bald and some are not, and
if baldness is uniquely defined by the number of hairs, then there
should be a transition point in between, and a single added hair is
bound to exceed it. We will discuss a list of reasons for our susceptibility
to the soritical reasoning in the conclusion. 

One reason, however, is central for this paper: we are adopting, as
a rule of the game, the assumption $\Sup$, but we do not want to
believe in its consequences. That is, if the function $\pi$ in Max's
rules exists, then a boundary point $v$ should exist too. If one does
not want to believe in the existence of such a point, then Max's rules should
be disbelieved too. Of these rules, M1 is merely a description
of clear-cut cases, and M2 and M3 are merely explicating the
meaning of being or not being close to 1. These rules are difficult
not to accept---provided one accepts the existence of a stimulus-effect function to begin with.


\section{Probabilistic Supervenience}\label{sec:prob}

\subsection{Inconsistent responses}

Zora is in almost all respects like Max: her set of stimuli is $S=[0,1]$,
and to every instance of $x$ she responds by saying $r_1$ or $r_0$.
But these responses pertain to the instances of $x$ rather than the
value of $x$. The reason for this is that Zora does not assign responses
to stimuli consistently. In the imagined multiverse with an infinity
of Zoras responding to $x$, generally, some responses will be $r_1$
and some $r_0$. There is, however, a well defined probability of
occurrences of $r_1$ in response to $x$, which we denote by $p(x)$.
Zora's rules parallel Max's, and are as follows:
\begin{enumerate}
\item[(Z1)] there are $x_0,x_1 \in S = [0,1]$ such that $p(x_0) = 0$ and $p(x_1) = 1$;
\item[(Z2)] the function $p(x)$ is (non-strictly) increasing.
\end{enumerate}
Strictly speaking, the stimulus-effect function $\pi(x)$ in this setting is the probability distribution
\[
\pi(x)=\left[\begin{array}{cc}
r_1 & r_0\\
p(x) & 1-p(x)
\end{array}\right],
\]
but since this is determined entirely by $p(x)$,
we can view $p$ as the stimulus-effect function instead.

\subsection{A rational response}

Can a soritical trap be set based on Zora's rules? Let us invite
Alex and Eubilides again.
\begin{longtabu} to \linewidth {rX}
Eubulides: &  I have just described to you Zora's rules. Please
accept them. Will you agree that $p(1)=1$ and $p(0)=0$?\\
Alex: &  Yes, it follows from Z1 and Z2.\\
Eubulides: &  Will you agree
that,\\
& \begin{tabu} to \linewidth {rX}
(E3) & if $x \in [0,1]$ and $\varepsilon>0$ is sufficiently small, then $p(x\pm\varepsilon) = p(x)$?
\end{tabu}\\
Alex: &  No. This may be true for some $x$, but cannot be true for
all $x$. If it were, then the function would have to be constant
on the entire interval $[0,1]$, which is impossible since
$p(1)\not=p(0)$.
\end{longtabu}
\noindent This short dialogue establishes that $p(x)$ is not tolerant,
i.e., the system does not satisfy $\Tol$, whence one cannot form a soritical sequence.

It would not help if one used a discretization of $p(x)$,
e.g., by defining a new probability function $\widetilde{p}$ by $\widetilde{p}(x)=1$ if $p(x)\geq\nicefrac{1}{2}$, and
$\widetilde{p}(x)=0$ otherwise. This would effectively reduce
Zora's rules to Max's \citep{Cargile-1969}. 

\subsection{Conclusion}

Let us formulate two informal hypotheses (or more correctly, guiding principles) about the world (or about behaviors
in the world). The first one is the hypothesis that lack of supervenience
means the system behaves probabilistically.\\

\noindent \begin{tabu} to \linewidth {rX}
($\sim\Sup\equiv\Prob$) & All empirical
systems that violate the assumption of supervenience behave probabilistically.
\end{tabu}\\

\noindent The precise meaning of this hypothesis is this: when responses from
set $R$ do not supervene on stimuli from set $S$, then there is
a sigma-algebra $\Sigma$ on $R$ and a function 
\[
\lambda:S\rightarrow M_{(R,\Sigma)},
\]
where $M_{(R,\Sigma)}$ is the set of all probability measures
on the measure space $(R,\Sigma)$. Thus, if Zora finds
out that her responses $r_0$ and $r_1$ are not determined
uniquely by points in $[0,1]$, then she knows that every
point of $[0,1]$ is mapped into a probability distribution
uniquely described by $p(x)$. Of course, supervenience is merely
a special case of probabilistic behavior, with $p(x)$
attaining only the values $0$ and $1$. In view of this the hypothesis can
also be formulated thus: all empirical systems behave probabilistically.

The second hypothesis is that
lack of supervenience is ubiquitous in all situations where one is
likely to construct a soritical trap.\\

\noindent \begin{tabu} to \linewidth {rX}
($\sim\Sup$) & In all empirical systems
where the assumption of supervenience is not accompanied by plausible
identifiability of non-tolerance points or a plausible explanation
of non-connectedness, the supervenience assumption is violated.
\end{tabu}\\

\noindent Because of the vague term `plausible', this hypothesis is not
a well-formed scientific statement. The only reason for stating it
here in this imperfect form is that we are not concerned with the
exact sphere of applicability of this hypothesis. Rather, we are interested
in the possibility of using this hypothesis in specific situations,
when the non-deterministic nature of a behavior (lack of the kind of supervenience enjoyed by Max) can be empirically demonstrated. The hypothesis
$\sim\Sup$ essentially says such situations are ubiquitous. Probabilistic
supervenience therefore should be viewed as the first and simplest way
of dissolving soritical traps. For a similar view in the literature,
see \cite{Hardin-1988}.

\section{Anticipating objections}\label{sec:obj}

\subsection{Probabilistic predicates versus truth values}

The use of probabilistic supervenience may appear to be just a variant
of the degrees-of-truth or fuzzy sets approach to sorites \citep{Black-1937, Edgington-1997}, according to which, for some predicates $\Pi^{*}$
defined for all $x\in S$, the statements `$\Pi^{*}(x)$'
have non-classical truth values: not just $\T$ or $\F$, but any
number between 0 and 1. Let us denote this number by $\TV[\Pi^{*}(x)]$.
Let us assume a correspondence between a stimulus-effect function
$\pi$ and a `real-world' predicate $\Pi^{*}$ can be established.
For instance, Zora's stimulus-effect function 
\[
p(x)=\Pr[\textnormal{Zora says $x$ is close to 1}]
\]
can be paired with the vague predicate 
\[
P^{*}(x)\equiv x\textnormal{ is close to 1}.
\]
This pairing is far from obvious, as we do not know what being
`in reality' close to 1 means, and we do not know if Zora's understanding
of this predicate accords with any normative rules (except for her own
rules, Z1 and Z2). If, however, we overlook this difficulty, is it a tenable view that the truth of $p(x)$ equalling $p$ is equivalent to $\TV[P^{*}(x)]$ equalling $p$?

The answer to this question is negative. To see this, consider the
following. For every $x,y$ within the domain $S$, we have
\[
(\T[p(x)=p]\textnormal{ and }\T[p(y)=q])\mbox{ iff }\T[p(x)=p\wedge p(y)=q].
\]
But (using \L ukasiewicz and Tarski's many-valued logic rules,
cf.~\cite{Hajek-2003}), if conjunction is understood in the weak sense then we have
\[
(\TV[P^{*}(x)]=p\textnormal{ and }\TV[P^{*}(y)]=q)\textnormal{ implies }\TV[P^{*}(x)\wedge P^{*}(y)]=\min\{ p,q\}, 
\]
whereas if conjunction is understood in the strong sense then
\[
(\TV[P^{*}(x)]=p\textnormal{ and }\TV[P^{*}(y)]=q)\textnormal{ implies }\TV[P^{*}(x)\wedge P^{*}(y)]=\max\{ 0,p+q-1\} .
\]
Another example: by the rules of classical calculus of propositions,
\[
\textnormal{if }\T[p(y)=q]\textnormal{ then }\T(p(x)=p\Rightarrow p(y)=q),
\]
irrespective of whether $\T[p(x)=p]$ or $\F[p(x)=p]$.
But
\[
\textnormal{if }(\TV[P^{*}(x)]=p\textnormal{ and }\TV[P^{*}(y)]=q)\textnormal{ then }\TV[p(x)\Rightarrow p(y)]=\min\{ 1,1-p+q\} .
\]

On the other hand, there is one useful parallel between the two approaches.
If $\overline{P}^{*}(x)$ is understood as the internal
negation of $P^{*}(x)$, and if we associate it with 
\[
1-p(x)=\Pr[\textnormal{Zora says $x$ is not close to 1}],
\]
we have
\[
\TV[P^{*}(x)]=p\textnormal{ iff }\TV[\overline{P}^{*}(x)]=1-p.
\]

\subsection{\label{sec:Higher-Order-Probabilities?}Higher-Order Probabilities?}

There is an equivocation in using the term `response'. When Max
or Zora respond to an instance of $x$ by $r_1$, as being `close to 1', the latter is
a response. But only for Max is it also a stimulus-effect, in the
sense of supervening on $x$. For Zora it is not: her stimulus-effect
is the probability $p(x)$ of saying $r_1$. This distinction
is the reason we use the term `stimulus-effect' as separate from
`response'. It is, however, always possible, whatever stimulus-effect is being considered, to view it as a response consistently
given to every instance of $x$. Thus, if Max instead of $r_0$
or $r_1$ responded to every instance of $x$ by saying `$r$',
where $r$ is some number between 0 and 1, it would be equivalent to
Zora saying $p(x)=r$. 

The question arises: if this is a possible point of view, could not
then the assignment of probability distributions to stimuli be subject
to probabilistic considerations of their own? Indeed, the new Max
who responds to every instance of $x$ by a number between 0 and 1
may be found to do this inconsistently, and then, by our hypothesis
$\sim\Sup\equiv\Prob$, he should be behaving probabilistically.
This would entail a probability distribution on $[0,1]$,
for every value of $x$. This distribution can be described by a distribution
function 
\[
A_x(r)=\Pr[\textnormal{(new) Max chooses a number}\leq r\textnormal{ for this instance of }x],
\]
for all $r\in[0,1]$ and $x\in[0,1]$. Could not the same reasoning apply to a new Zora whose probabilities
of responding $r_1$ are not consistent? This would mean the following analogue of the distribution function above:
\begin{align*}
	Z_x(r) = & \Pr[\textnormal{(new) Zora responds to an instance of }x\textnormal{ by saying }r_1\\
	& \textnormal{ with a probability}\leq r],
\end{align*}
for all $r\in[0,1]$ and $x\in[0,1]$.

Here, however, the analogy ends. Max and new Max do respond very
differently, but new Zora is merely Zora with a different probability
function $p(x)$. Indeed, probability $p(x)$
changing in accordance with $Z_x(r)$ is merely a new
probability function
\[
\widetilde{p}(x)=\int r\dd Z_x(r),
\]
which is the expected value of $r$ distributed in accordance with
$Z_x(r)$. The probability of Zora saying $r_1$ in
response to $x$, if it changes probabilistically, is merely another
probability of Zora saying $r_1$ in response to $x$. In other
words, under our hypotheses the probabilities of observable responses
always supervene on stimuli. We never need probabilities of probabilities.
The general statement is
\begin{thm*}[Woodbury-Savage Reduction]
 A probability distribution of probability measures on a measure
space $(R,\Sigma)$ is equivalent to a measure on the measure
space $(R,\Sigma)$.
\end{thm*}
This version is a trivial generalization of the theorem given in \cite{Savage-1972}. The equivalence is understood in the sense of implying one
and the same probability with which $r\in R$ falls within every measurable
subset $E$ of $R$. The proof obtains by denoting the `second-order'
probability measure $\mu(\lambda)$, where $\lambda$ is
an ordinary (`first-order') probability measure on $(R,\Sigma)$,
and observing that 
\[
\Pr[r\in E]=\int\lambda(E)\dd\mu(\lambda)
\]
(in the Lebesgue sense). One can easily check that this probability,
taken over all $E\in\Sigma$, is a well-formed probability measure
on $(R,\Sigma)$.

\medskip
\noindent \textbf{Example.} Suppose we are handed one
of two coins to flip, Coin 1 with $\Pr[\textnormal{head}]= p$
and Coin 2 with $\Pr[\textnormal{head}]=q$, and
suppose we are handed Coin 1 with probability $r$. Then the
probability with which the outcome of a given toss will be `head' will be
the same as that of flipping a single coin having 
\[
\Pr[\textnormal{head}]=p \times r + q\times (1-r).
\]
After all, a single fair coin ($\Pr[\textnormal{head}]=0.5$)
can very well be conceptualized in this way with $p = 1$, $q = 0$, and $r = 0.5$ (i.e., both sides of Coin 1 are heads, both sides of Coin 2 are tails, and we are handed Coin 1 exactly half the time).

\medskip
The reduction argument, mutatis mutandis, has been repeated in the
philosophical literature numerously \citep{Kyburg-2013, Pearl-2013}.
The opinion in favor of `higher-order' distributions \citep[see, e.g.,][]{Hansson-2008} is based on a logical confusion. If the alternation
of Coin 1 and Coin 2 in our example is arranged in a series,
e.g., each occurrence being tied to a particular point in time (or according to any other way of tagging individual
occurrences), then the pattern of changes in time can be detected,
or at least theoretically considered. This change of probabilities
in time, however, is not a probability distribution of probabilities.
One can speak of such a distribution at any given moment, but then
by the reasoning above it can always be replaced by a single probability.
One will have therefore a probability, say Zora's $p(x)$,
developing in time, i.e., treated as a function $p(x,t)$ giving
the probability of $r_1$ as a function of $x$ and $t$. In this
case, $t$ should simply be included within the description of the
stimuli $x$ (i.e., the domain of the system should properly consist of pairs $(x,t)$).

\section{Rule-making}\label{sec:rules}

\subsection{Normative rules}

We turn now to how one sets up normative rules that make the behavior
of anyone following these rules completely predictable. Max's rules,
e.g., do not fall in this category. They are under-definitive in
the following sense: while they compel anyone following them to construct
the stimulus-effect function $\pi$ in a particular way ($r_1$
up to some point $v$, $r_2$ thereafter), they do not determine
this function uniquely. Zora's rules are under-definitive with respect
to probabilities because they allow different
people to use different functions $p(x)$, requiring only
that these functions be monotonic and attain the value $0$ and 1
at, respectively, $x=0$ and $x=1$. 

For simplicity, we will focus on Max's rules in our discussion
of rule-making. Let Max, Alex, and Zora come together to determine
a definitive form of these rules. They know that the stimulus
effect function is 
\[
\pi(x)=
\begin{cases}
r_0 & \text{if } x<v\\
r & \text{if } x=v\\
r_1 & \text{if } x>v
\end{cases}
\]
where $v$ can be any number in $[0,1]$, and $r$ is either
$r_0$ or $r_1$ (constrained by the stipulations $r=r_0$ if
$v=0$, and $r=r_1$ if $v=1$). How do they make the choices?

\subsection{Arbitrariness}

We begin by observing that the procedure of deciding on $v$ and $r$
can very well be viewed as behavior, albeit very specific behavior. There is
a single stimulus here, the interval $[0,1]$ to be partitioned,
and for something to supervene on it simply means this something should
be uniquely determined. In the real world, if one were to investigate rule-making behavior of this kind one would have to form many
similar triads of people and observe their decisions. In our multiverse
picture, we think of an infinity of worlds with Max-Alex-Zora
triads making these decisions. Their decision is represented by the
pair $(v,r)$. Since there is nothing known to us that
would compel all these triads of rule-makers to make the same choice,
we can invoke our hypothesis $\sim\Sup \equiv \Prob$ to assert that $(v,r)$
is chosen probabilistically. This means that there is a \emph{single} probability
distribution of the $(v,r)$s, and therefore that 
this distribution supervenes on the single stimulus $[0,1]$. (We know
that this solution is stable: any probability distribution of the
probability distributions of the $(v,r)$s is equivalent to
a single probability distribution of the $(v,r)$s.)

It may, however, be more interesting for a philosopher to consider
what decision Max, Alex, and Zora \emph{ought} to make rather than what
decisions they do make. Is there some way of determining, based on Max's rules alone, what the choice
of $(v,r)$ ought to be? The obvious
answer is no: the choice is entirely arbitrary. There is nothing in
the rules known to Max, Alex, and Zora that would make, say, $(1/2,r_1)$ a better choice than any other choice of $(v,r)$, let alone the \emph{only} possible choice. Max,
Alex, and Zora are in the position of Buridan's ass surrounded by an
infinity of identical hay stacks. We submit that the arbitrariness of the choices involved may be one of
the main reasons for the uncanny persuasiveness of sorites.

\subsection{Justification}

It seems
likely to us that people who erroneously accept the universality of
the soritical step in the classical soritical traps with heaps and
baldness may do so because they are correctly aware of their inability
to justify any precise rules about baldness and heaps. People find
it difficult (perhaps, impossible) to make choices arbitrarily. People
want justifications, and when they do not have any they cast lots
and consult spirits. Alex, in her conversations with Eubulides, may
very well understand that she cannot accept the universality of the
soritical step because she knows that a boundary $v$ must exist.
Max, Alex, and Zora together can make the rule that $v$ is to be
set to $3/4$, yet not be aware of any principle
(law of nature, convention) to justify this rule or
to prevent setting $v$ to $3001/4000$.
They realize they are unlikely to find a foundation for their rule
that would not itself be equally unfounded, and as a result
they correctly think $v$ \emph{could} very well be changed to $3001/4000$, or, for that matter, to $1/8$. The smallness of the change is not significant, and serves merely to remind them that their rule must be precise. 

\subsection{Correctness}

Arbitrariness of choices means also they cannot be wrong or right. Epistemicists
\citep{Keefe-2000, Sorensen-1988a,Williamson-1994, Williamson-1997, Williamson-2000} disagree with this: they seem to consider the task of setting a $v$
between $0$ and $1$ as a discovery of something that objectively exists,
and even uniquely exists. In other words, for them there must be a `correct' boundary $(v,r)$
within the interval $[0,1]$ between the numbers close
to 1 and those not close to 1. We do not see why this should be the case (joining in this respect
other authors, e.g., \cite{Gomez-torrente-1997}, \cite{Tye-1997}). As mentioned in Section \ref{sec:det}, in cases where responses supervene on stimuli, we do not see why one cannot learn (or approximate) the position of a true boundary. But in cases where, somehow, we fundamentally cannot know the value of a true boundary, we do not see how an objective `correctness' of such a boundary can be justified. Is this not like arguing that because a newborn child will eventually have a name, and because this is a fact about the future and as such holds even \emph{before} the baby is named, there should already at that point be an objectively right or wrong name that the baby ought to be given? 

There is, however, one aspect about which the epistemic position seems
to be indisputable. Even if $v$ is known to us with arbitrary precision
but not precisely, we can never learn whether $\pi(v)$
is $r_1$ or $r_0$. 
The knowledge of $\pi(v)$ is
conditioned upon the knowledge of $v$, and the latter cannot be achieved
by observations (even by the idealized observations of an infinity of
Maxes in parallel worlds). The question to ask here is whether
this determination matters for predicting or understanding Max's
behavior. If the difference between $\pi(v)=r_0$ and
$\pi(v)=r_1$ has observable consequences, then the true
choice can be made based on them. But this will lead us outside Max's rules and the description of $x$ as the only stimulus given in this task. 

Supervaluationists (\cite{Dummett-1975}; \cite{Fine-1975}; \cite[Chapters
7,8]{Keefe-2000}) accept the fact of arbitrariness. The Max example is very similar
to Kit Fine's example with nice$_1$: if we rename `close to 1'
into `nice$_1$', then we know that 1 is `nice$_1$',
that 0 is `not nice$_1$', and all the numbers in between can
be labeled by `nice$_1$' and `not nice$_1$' arbitrarily
(within the constraints of rules M2 and M3). We cannot, however,
see a reason for the supervaluationist insistence on considering all
possible labelings. From a logical point of view, this is the only
correct way of looking at the situation if the goal of looking at
it is to find out propositions that preserve their truth value under
all possible choices. But one can be equally interested in propositions
that are true under some choices of labelings, or even under one specific
such choice.

\section{Conclusion: Why is sorites psychologically persuasive?}

\subsection{Summary}

It seems that the classificatory sorites is not a very complex issue.
When faced with a soritical trap, one has first to examine the assumption
of supervenience. If it holds, then either tolerance or connectedness
have to be rejected. If supervenience does not hold, then a soritical
trap cannot be formulated. But one can assume then that the assignments
are probabilistic and supervenience applies to the probabilities.
A soritical trap then cannot be formulated as in the first case. If one has to make
a deterministic rule, one is faced with the necessity of making arbitrary
choices. These choices are unjustifiable (otherwise they would
not be arbitrary), but unavoidable and rational.

\subsection{Theorizing about sorites}

Why is then sorites is considered such a very hard problem (\cite{Priest-2004}; \cite{Varzi-2003})? 

Note that no soritical traps exist for the hypothetical Max or Zora
who answer questions like `is this number close to 1, yes or
no'? The supervening effect of the responses in their respective cases (deterministic and probabilistic), and
the boundary $v$ in Max's, can be determined. The trap only
exists for someone who, like Alex, theorizes about the performance of
Max and Zora. In effect, Alex is supposed to construct a theory of
sorites in her mind and explicate all the assumptions involved. This
is indeed not trivial. (At least not for us, trying to do
so in this paper).

We have pointed to two main sources of difficulty in theorizing about
sorites. The first is that deterministic supervenience is implicitly assumed
in the formulation of a soritical trap, but Alex is asked to rely on
her intuition concerning notions which real behavior does not supervene on---it only behaves probabilistically. And this is not just human
behavior applied to vague notions. Borrowing an example from \cite{DD-2010a}, a big rusty two-pan balance with a fixed weight in
the left pan and a variable weight $x$ in the right (the stimulus, in the behavioral approach) may be at
equilibrium or it may tip right or left (the response). Can the addition
or deletion of a single atom upset the balance, causing it to tip?
One's intuition revolts against the idea of something big and clumsy
being sensitive to microscopic changes, but the revolt is likely to
be pacified if one realizes that as the balance approaches the state
of unstable equilibrium its behavior must become probabilistic. Can
the probability of the balance tipping to the right increase as a
result of adding a single atom to $x$? Yes, of course, by a very
small amount. The reduction theorem of Section \ref{sec:Higher-Order-Probabilities?}
and the fact that deterministic behavior is merely a special case
of probabilistic make the probabilistic dissolution of sorites both
firm and universally applicable. 

The second difficulty with sorites pointed out in this paper is that a theorist who
tries to make a rule relating something like the notion `bald'
to hypothetical stimuli (the number of hairs) is thinking about
the justifiability of the possible rules when, in fact, there are none
as the situation is truly arbitrary. A logical fallacy is then committed,
as the lack of justification for specifying
any given boundary is being mistaken for the impossibility
of doing so. If our task is to send three different
postcards to Max, Alex, and Zora, but our instructions do not specify
whom to send which postcard to, the rational behavior is
to arbitrarily choose among the six possible versions. A `correct' choice
does not exist. If one considers rule-making as special behavior,
then our identical copies in the parallel worlds should send
the postcard in all six different ways (with possibly unequal probabilities, indicating various biases on our part). 

These two reasons for soritical persuasiveness are definitely not
the only ones. There are purely psychological reasons one may commit
logical fallacies on account of. Thus, \cite{Williamson-1997} correctly points out that
a person asked to judge the truth of `If a head is not bald, then
removing one hair would not make it bald' may replace the antecedent
with `If a head has lots of hairs', confusing the notion of `not
bald' with that of `typical person who is not bald'. But we do not wish
to get into psychological reasons like this.

\subsection{Classificatory versus comparative sorites}

We do, however, wish to address another possible reason, namely,
the logical confusion of the classificatory sorites with the comparative
one. The formal difference between the two is the following. In the classificatory sorites we have an arbitrary set of stimuli $S$, an arbitrary set of stimulus-effects $R$, and the function $\pi$ mapping $S$ into $R$. A classificatory
soritical sequence $x_1,\ldots,x_n$ cannot exist in classical
logic because it is contradictory: if $\pi(x_i)=\pi(x_{i+1})$
for all $i=1,\ldots,n-1$, then it is impossible to have $\pi(x_1)\not=\pi(x_n)$.
In the comparative sorites we have pairs of stimuli from $S\times S$
and only two (fixed) possible responses, `same' and `different'. Assuming
the supervenience of these responses on the pairs of stimuli, a comparative soritical
sequence is $x_1,\ldots,x_n$ such that $(x_i,x_{i+1})$ are mapped to `same' for all $i=1,\ldots,n-1$, yet $(x_1,x_n)$ is mapped to `different'.
Here, unlike in the classificatory sorites, there is no \emph{logical} contradiction, and the possibility of a soritical
sequence depends on the definition of `same' and `different'. If one defines
two numbers to be `same' if they differ by no more than $0.5$, and `different' otherwise,
then a comparative soritical sequence can be readily constructed (say,
$x_1=0$, $x_2=0.4$, $x_3=0.8$). If one defines two numbers
(between 0 and 1) to be `same' just when they have the same first
5 digits in their decimal expansion, then a comparative soritical
sequence does not exist. 

The confusion in question occurs when one explains the necessity of
the classificatory soritical step by the fact that one cannot distinguish
two sufficiently similar stimuli. But this does not apply to the
classical sorites involving baldness and heap, nor to our Max and
Zora examples. A number $x$ is assumed to be known to the respondents
(and to Alex theorizing about the responders) precisely, and the equality
$x=y$ is understood as precise equality, not approximate one. 

The
explanation through comparative sorites could plausibly work in what
is called `observational' sorites: e.g., if Aliya is asked to
judge whether a given color patch is `red' or `not red', she
may be thought (by Alex, theorizing about her own performance) to be unable
to tell apart two very similar shades of color---so that however
she might understand `red', she will have to give the same response to
both these shades of color. Even for the observational situations, however, the explanation in
question is dubious. It hinges on a specific understanding of the
comparative sorites for which we do not have any empirical evidence
(and observational sorites is, of course, about empirical situations).
In real human behavior (or the behavior of a technical gadget), responses
like `same' and `different' do not supervene on stimulus pairs
if they involve very close stimuli. It is a fundamental empirical
fact that sometimes people (or gadgets) will judge $(x,x)$
as `different' and $(x,x+\varepsilon)$ as `same'.
One cannot simultaneously eliminate `errors' of these two types. If
one takes into account the probabilistic nature of supervenience
here and computes the matching relations between stimuli as characteristics
of the probability distributions, comparative soritical sequences
become less than obvious. Carefully collected experimental evidence seems to be
in favor of the hypothesis of \cite{DD-2010b} that comparative
soritical sequences do not exist \citep{DP-2010,DP-TP,DC-2006}.

\bibliography{Sorites}
\bibliographystyle{apalike}

\end{document}